\documentclass[runningheads,a4paper]{llncs}

\usepackage{amssymb}
\usepackage{graphicx}

\input{mystyles.x}
\usepackage{contour}

\begin{document}
\contourlength{0.2pt}
\contournumber{50}
\mainmatter  % start of an individual contribution

\title{Inference of Tissue Haemoglobin Concentration From Stereo RGB}

\titlerunning{Inference of Haemoglobin Concentration From Stereo RGB}

\author{\doubleblind{Geoffrey Jones$^1$\and Neil T. Clancy$^{2,3}$\and Yusuf Helo\and Simon Arridge$^1$\and Daniel S. Elson$^{2,3}$\and Danail Stoyanov$^1$}}
\authorrunning{\doubleblind{G. Jones et al.}}

\institute{\doubleblind{$^1$Centre for Medical Image Computing, University College London\\$^2$The Hamlyn Centre, Institute of Global Health Innovation, Imperial College London\\
$^3$Department of Surgery and Cancer, Imperial College London}}%\mailsa\\
%\url{http://www.springer.com/lncs}}


\toctitle{Lecture Notes in Computer Science}
%\tocauthor{Authors' Instructions}
\maketitle

\vspace{-2em}
\begin{abstract}

Multispectral imaging (MSI) can provide information about tissue oxygenation, perfusion and potentially function during surgery. In this paper we present a novel, near real-time  technique for intrinsic measurements of total haemoglobin (\THb) and blood oxygenation (\SatO) in tissue using only RGB images from a stereo laparoscope. The high degree of spectral overlap between channels makes inference of haemoglobin concentration challenging, non-linear and under constrained. We decompose the problem into two constrained linear sub-problems and show that with Tikhonov regularisation the estimation significantly improves, giving robust estimation of the \THb . We demonstrate by using the co-registered stereo image data from two cameras it is possible to get robust \SatO\ estimation as well. Our method is closed from, providing computational efficiency even with multiple cameras. The method we present requires only spectral response calibration of each camera, without modification of existing laparoscopic imaging hardware.
We validate our technique on synthetic data from Monte Carlo simulation
% of light transport through soft tissue containing submerged blood vessels
and further, {\em in vivo}, on a multispectral porcine data set. %with a hardware acquired multispectral data set for comparison.
\vspace{-1em}
\end{abstract}

\section{Introduction}

\begin{figure}[t]
\begin{minipage}{\linewidth}\centering
\includegraphics[width=\textwidth]{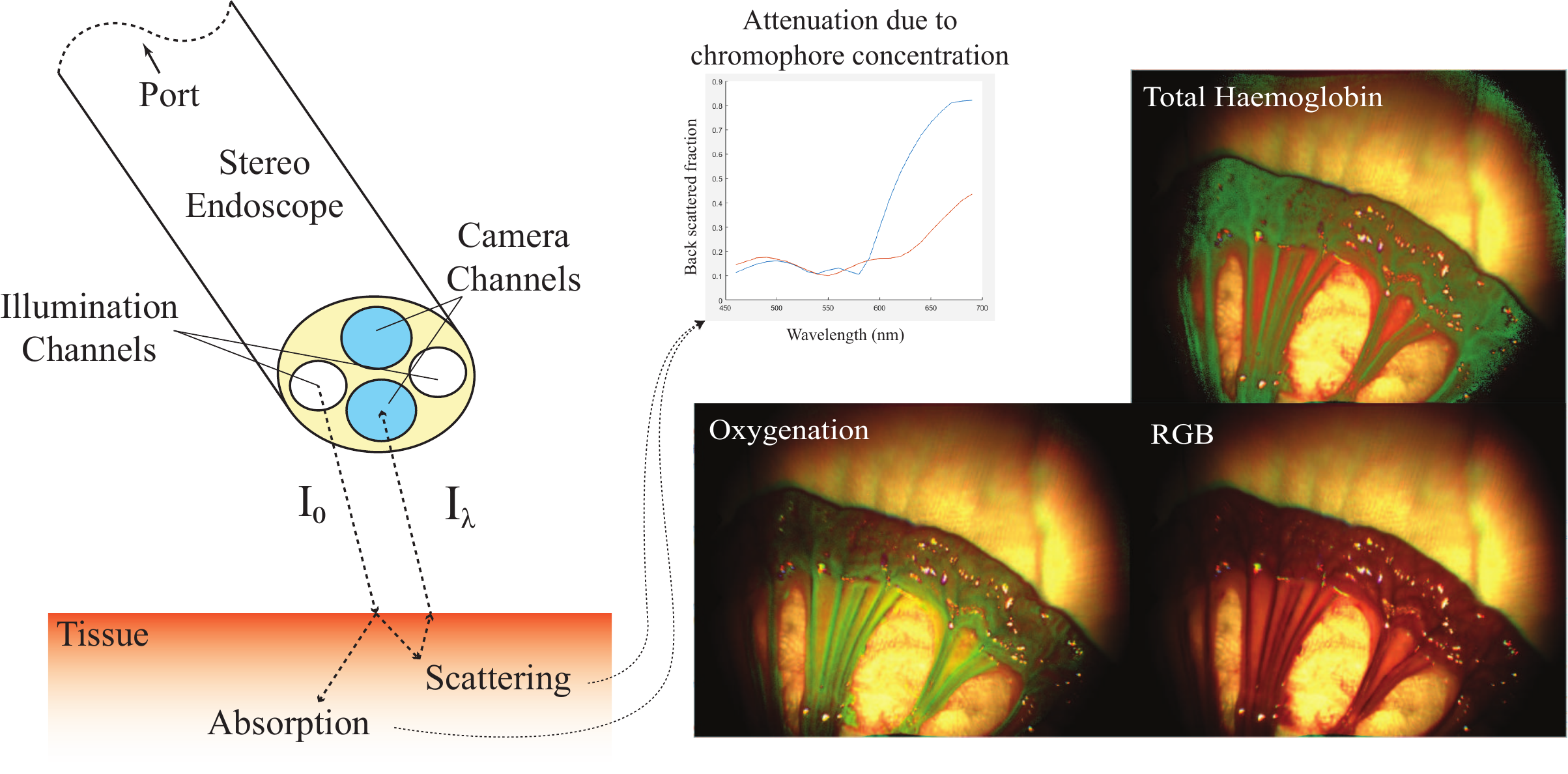}
\end{minipage}\vspace{-.5em}
\caption{A stereo RGB endoscope acquires images of the tissue surface under white light illumination. Inside the tissue light is attenuated due to the process of scattering and absorption to a greater, or lesser, degree depending on wavelength and concentration of \HbO\ \Hb . These concentrations can be calculated and presented to the surgeon in the form of information overlays that map \THb\ or \SatO\ within tissue.\label{f:set up}}
\vspace{-1.5em}\end{figure}

Intraoperative imaging is critical for guiding surgical procedures, especially in minimally invasive surgery (MIS) where the surgeons' access to the surgical site is indirect and restricted \cite{darzi2002recent}. Current white light imaging is mostly limited to providing information from tissue surfaces and does not help the surgeon to identify structures within the tissue such as blood vessels. Laparoscopic images contain only macroscopic structural and radiometric information, that does not directly highlight tissue function or characteristics which may be used to identify malignancy. Pathological signals such as oxy and de-oxy haemoglobin (\HbO, \Hb) concentration, often correspond to tissue structure \cite{sorg2005hyperspectral} or viability \cite{Clancy2014transplant} and are detectable by their characteristic attenuation of light in the visible wavelength range. Detecting and displaying this information {\em in vivo} could provide a powerful tool to the surgeon, but current imaging solutions often demand modification the laparoscopic imaging hardware or protocols.

MSI is an attractive modality for intraoperative surgical imaging because it is non-ionising and compatible with laparoscopic instrumentation.
It can be used to measure ischeamia {\em in situ} \cite{NighswanderRempel2002tissueOx} and in MIS to measure the oxygenation of tissue or to identify malignant tissue, where the increased vascularisation causes a local increase in \THb\ \cite{Claridge2007}. Bowel perfusion assessment \cite{Clancy2015:BowelOx} and uterine viability post-transplantation \cite{Clancy2014transplant} have been achieved by measuring the oxygenation saturation and total haemoglobin in the transplanted organ. Central to these type of techniques is a liquid crystal tunable filter (LCTF) which is used to serial capture band limited images, giving a high spectral resolution with the trade off of blur and misalignment when imaging dynamic tissue \cite{Clancy12}. Rapid filtering with maximally discriminative filter set can still enable estimation of the haemoglobin concentration from fewer measurement, but requires hardware modifications \cite{Wickert:2014}. Fast capture techniques directly utilising RGB images are possible via use of Wiener filtering to estimate the latent multispectral information \cite{Nishidate2013}. Hybrid approaches using several multi bandpass filters can capture full multispectral data at high frame rates \cite{Fawzy2015}. This technique tailors the filters to a specific RGB sensor so would require a break in surgery to switch imaging hardware. Temporal analysis of tissue using RGB video can also be used for estimation of oxygen saturation \cite{Guazzi2015}; however this is not an instantaneous approach, requiring sufficient time to detect periodic pathological processes.

With this paper, we develop a method for estimating \THb\ and \SatO\ by using the RGB sensors in stereo laparoscopes, which are already the prime imaging modality robotic MIS. The calibrated sensors' response curves define the mapping of the latent multispectral into RGB space we invert this process using a Tikhonov regularisation scheme to preserve smoothness. This reduces the problem into a two step process with the first of step having a closed form solution, enabling rapid processing of full frame stereo data.
We validate our technique using synthetic data generated from Monte Carlo (MC) simulation to evaluate the robustness to sensor measurement noise.
%framework \cite{MMCFang2010}measured optical characteristics of blood \cite{Bosschaart2013} and sub mucosa \cite{Bashkatov2014} input into the .
We further validate the method on {\em in vivo} data by using the multispectral derived result \cite{Clancy12} as the ground truth, showing our method is a close approximation of full MSI analysis. Our results are promising and suggest that it may be possible to provide additional information during surgery simply by resolving the existing imaging signal.\vspace{-.5em}

\section{Method}\vspace{-.5em}
Given a multispectral measurement comprising many non-overlapping band limited individual measurements $I_\lambda$, and a corresponding vector of initial illumination $I_{0,\lambda}$, the estimation of the concentration parameters ${\bm\alpha}$ is a straight forward least squares fitting. Where $\lambda$ corresponds to the the wavelengths of the band centres. This result is simply from rearranging the Beer-Lambert equation as such for an individual wavelength:
\begin{equation}
 -log(\frac{I_\lambda}{I_{0,\lambda}}) = {\bm \xi}{\bm\alpha}
\end{equation}
The attenuation coefficients ${\bm\xi}$ for the chromophores of \Hb\ and \HbO\ are dependant on scattering and absorption characteristics which are often given in the context of transmission. For computational efficiency, and to convert this to a backscattering context, we composite these into a single attenuation factor (Fig. \ref{f:set up}) determined via MC simulation \cite{Dunn2005}. To apply the Beer-Lambert equation the multispectral bands must be non-overlapping and each very narrow, given these conditions then this expression can be solved using a non-negative least squares solver to ensure positivity: we utilise the fast non-negative least squares method of \cite{Bro1997} to constrain the estimation from multispectral data.

In our case the measurement is actually from two cameras yielding 6 channels each with significant spectral overlap and spanning a wide spectral range. In order to preserve the conditions of the Beer-Lambert equation it would be necessary to pose the solution as the minimisation of 6 variably weighted sums of exponential terms. Instead of a direct non-linear approach we solve this in two steps, initially we estimate the latent multispectral data and then from this perform the standard least squares fitting to then estimate the chromophore concentrations.  
 
\begin{figure}[t]\hspace{-2.5em}
\begin{minipage}{0.64\linewidth}\centering
\includegraphics[width=\textwidth]{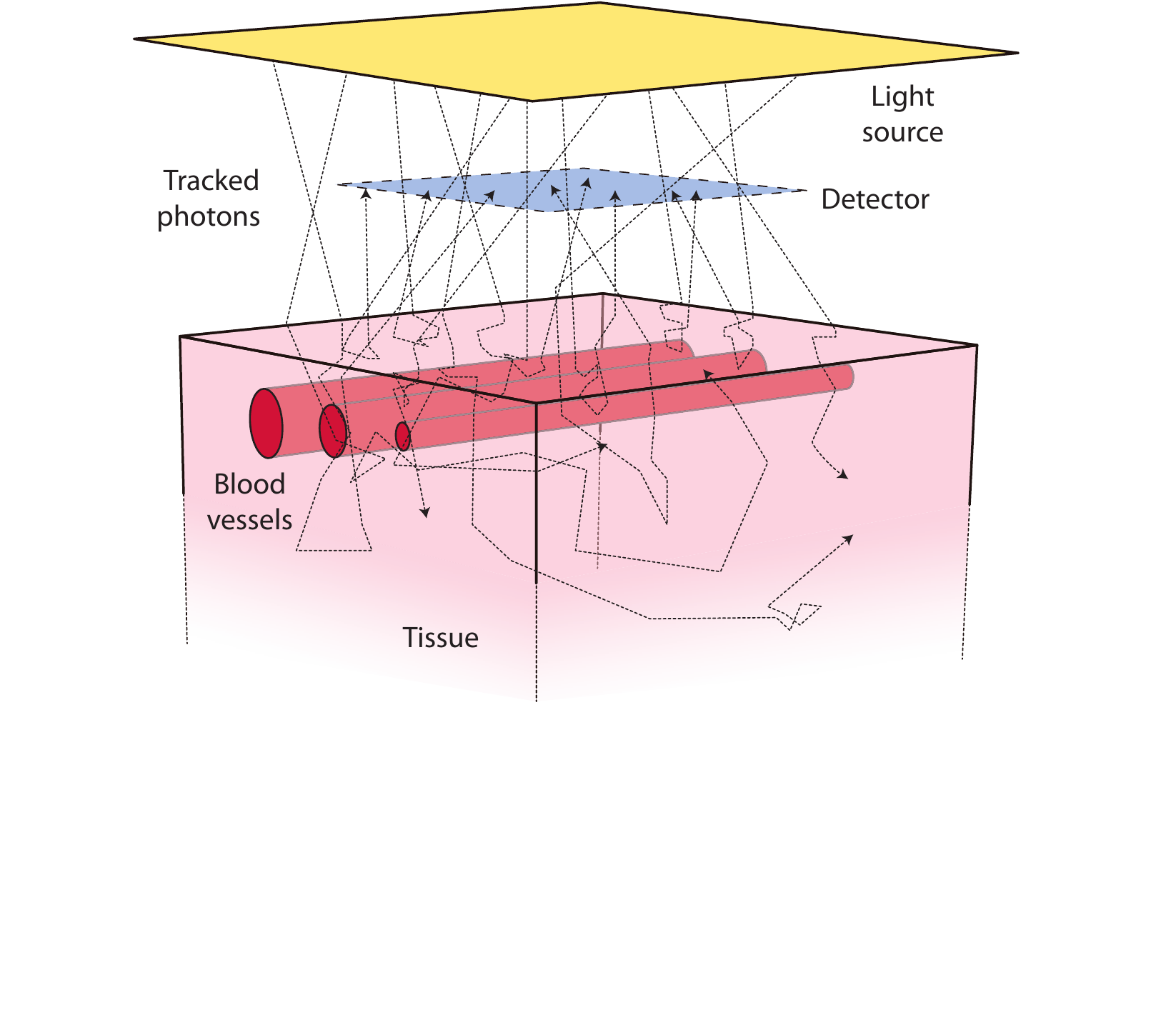}\vspace{-6.em}
\end{minipage}\hspace{-1.em}
\begin{minipage}{0.43\linewidth}\centering%\vspace{-5.em}
\includegraphics[width=\textwidth]{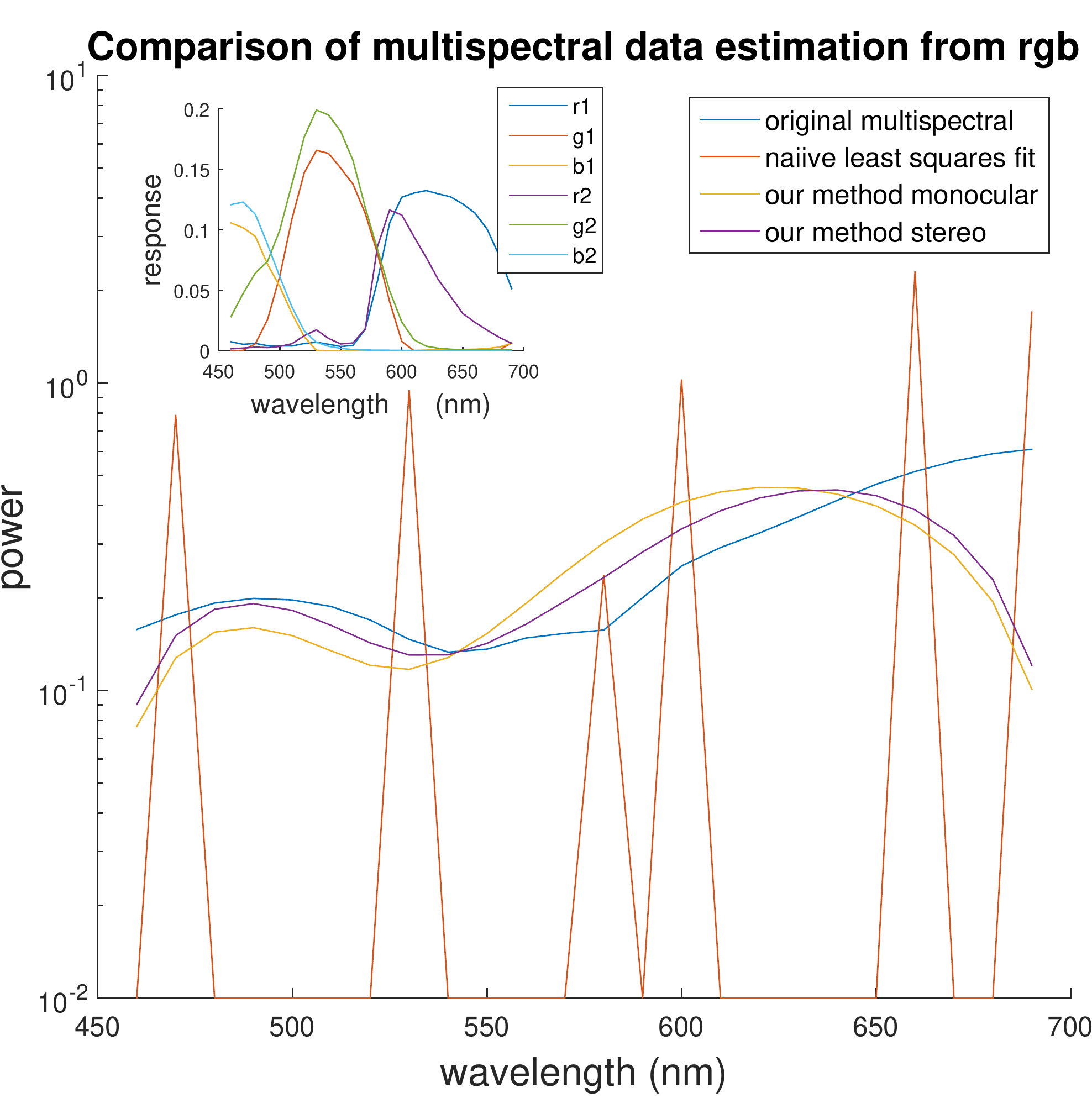}
\end{minipage}\vspace{-1.em} \caption{L) The synthetic model used for Monte Carlo simulation of test data comprising variable diameter blood vessels embedded within soft tissue. R) Three estimates of the multispectral data from RGB. Stereo sensor response inset. \label{f:setup}\label{f:specest}}\vspace{-1.8em}
\end{figure}

The na\"{\i}ve approach of estimating the multispectral data from the measured RGB image data $I_{RGBs}$ is to invert the spectral response curve of the camera $C$ solving the linear system though a least squares minimization such as:
\begin{equation}
I_\lambda  = \argmin_{I_\lambda} \left\| C I_\lambda - I_{RGBs} \right\|^2
\end{equation}
 where $\left||\cdot|\right|$ is the $L^2$ norm. This yields a poor estimation of the true multispectral data due to the problem being vastly unconstrained. $I_\lambda$ can have an order of magnitude more entries then $I_{RGBs}$, thus the na\"{\i}ve solution, while correct, is often a metamerism of the true multispectral data for the given $C$, as shown in Fig \ref{f:specest}. To constrain the estimation we impose a prior ($\Gamma$) on $I_\lambda$, this is imposed using a Tikhonov regularization:
\begin{equation}
I_\lambda  = \argmin_{I_\lambda} \left\| C I_\lambda - I_{RGBs} \right\|^2 + \left\|\gamma \Gamma I_\lambda \right\|^2 \label{e:regtik}
\end{equation}
The strength of the prior is regulated by the scalar $\gamma = 0.01$. This is typically solved implicitly as:
\begin{equation}
I_\lambda = (C^{T}C+ \Gamma^{T} \Gamma )^{-1}C^{T}I_{RGBs}\label{e:solve}
\end{equation}
where $\Gamma$ is often the identity matrix thus minimising the overall size of $ I_\lambda$. However we use a Laplacian matrix for $\Gamma$ to penalise non-smooth $I_\lambda$, the Laplacian matrix is formed of ones on the leading diagonal and negative half on the first super and sub diagonals. This choice of $\Gamma$ is made because we expect the multispectral data to be similar to that predicted by the Beer-Lambert relationship,
%\begin{equation}
%I_\lambda \approx I_{0,\lambda} e^{\left(-\bm\xi\bm\alpha\right)}
%\end{equation}
which is mostly smooth across the visible wavelength range for $\bm\xi$ comprising attenuation due to oxy and de-oxy haemoglobin.

Our method can also be applied to a monocular imaging context by reducing the number of columns in $C$ and the length of $I_{RGBs}$. This allows concentration estimation to either happen jointly utilising data from 2 or more cameras or independently for each camera. 

%\textcolor{red}{NB:} currently the performance of this is at around 0.5MP/s so not really real time for full frame HD video. The main reason for this is the need to solve a linear system of dimension 24 inside the inner loop of all iterations i.e. $A_{24\text{x}24}x_{24\text{x}1} = b_{24\text{x}1}$. I'm doing this at the moment using a Cholesky decomposition. I have been trying to find a more efficient method that maybe holds onto a decomposition and then just updates the decomposition at each iteration instead of recomputing it. I think this may be possible as the update to the A matrix is small positive and along the diagonal only, thus the  matrix continues to be self-adjoint positive-semi-definite. The tricky thing however is that the A without the update is always near (as good as) singular and the update is not just a scalar multiple of the identity, other wise I could use an eigen decomposition $A = U DU^\text{T}$ to then compute the updated decomposition $A+\Gamma = U (D+\Gamma) U^\text{T}$

\section{Experiments and results}\vspace{-.5em}
\subsection{Camera Spectral Response Dependency}
Our method is dependant on having available an accurate spectral response calibration for the camera sensor(s), protocols do exist for capturing very high quality calibration with negligible error using a monochromatic light source \cite{EMVA1288}. In the context of MIS such lengthy and refined calibration is unlikely to be available and calibration will probably be performed via imaging coloured patches of known reflectance \cite{Hardeberg1998}. This type of calibration is less accurate and at high sensor noise levels during calibration results in erroneous spectral response curve. The effect of a miss calibrated spectral response acquired in this manner is to bias the result of the optimisation towards deoxy haemoglobin however the \THb\ measure appears to be moderately robust to sensor miss-calibration.

\subsection{Synthetic evaluation}
To create synthetic test data we simulated multispectral image data which was subsequently filtered to generate RGB camera responses. The optical characteristics of blood and colonic submucosa (soft tissue) were compiled from \cite{Bosschaart2013} and \cite{Bashkatov2014} respectively.
The synthetic phantom model comprised a homogeneous block of soft tissue with three superficial vessels containing either oxygenated or de-oxygenated blood. The three blood vessels had different uniform diameters of 2mm, 1mm and 0.5mm and the top edge of each vessel was at the same depth below the surface of the tissue at a depth of 1mm.

\begin{figure}[t]\vspace{-1.5em}\hspace{-.5em}
\begin{minipage}{0.7\linewidth}\centering
\includegraphics[width=\textwidth]{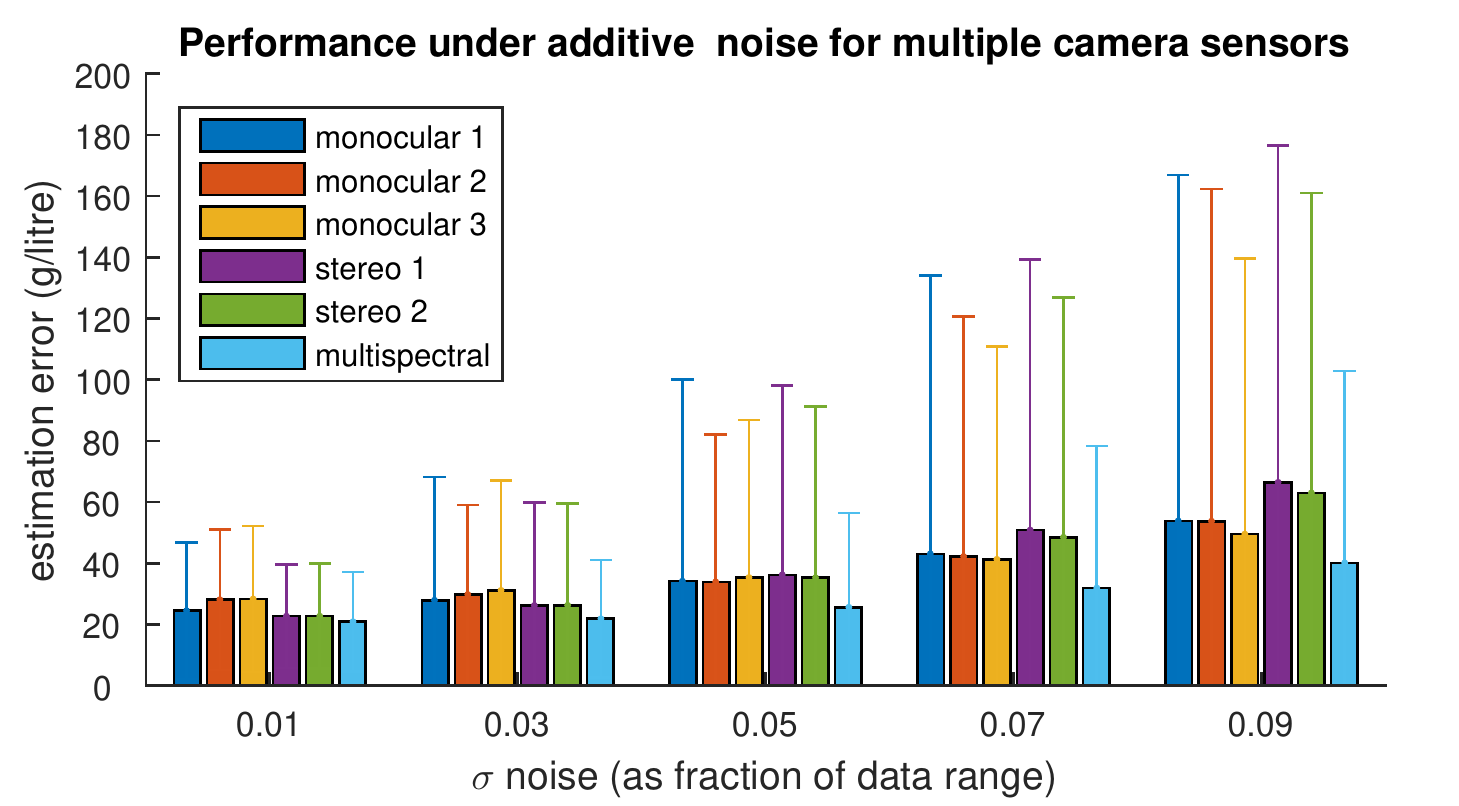}
\end{minipage}
\begin{minipage}{0.26\linewidth}\centering
\includegraphics[width=\textwidth]{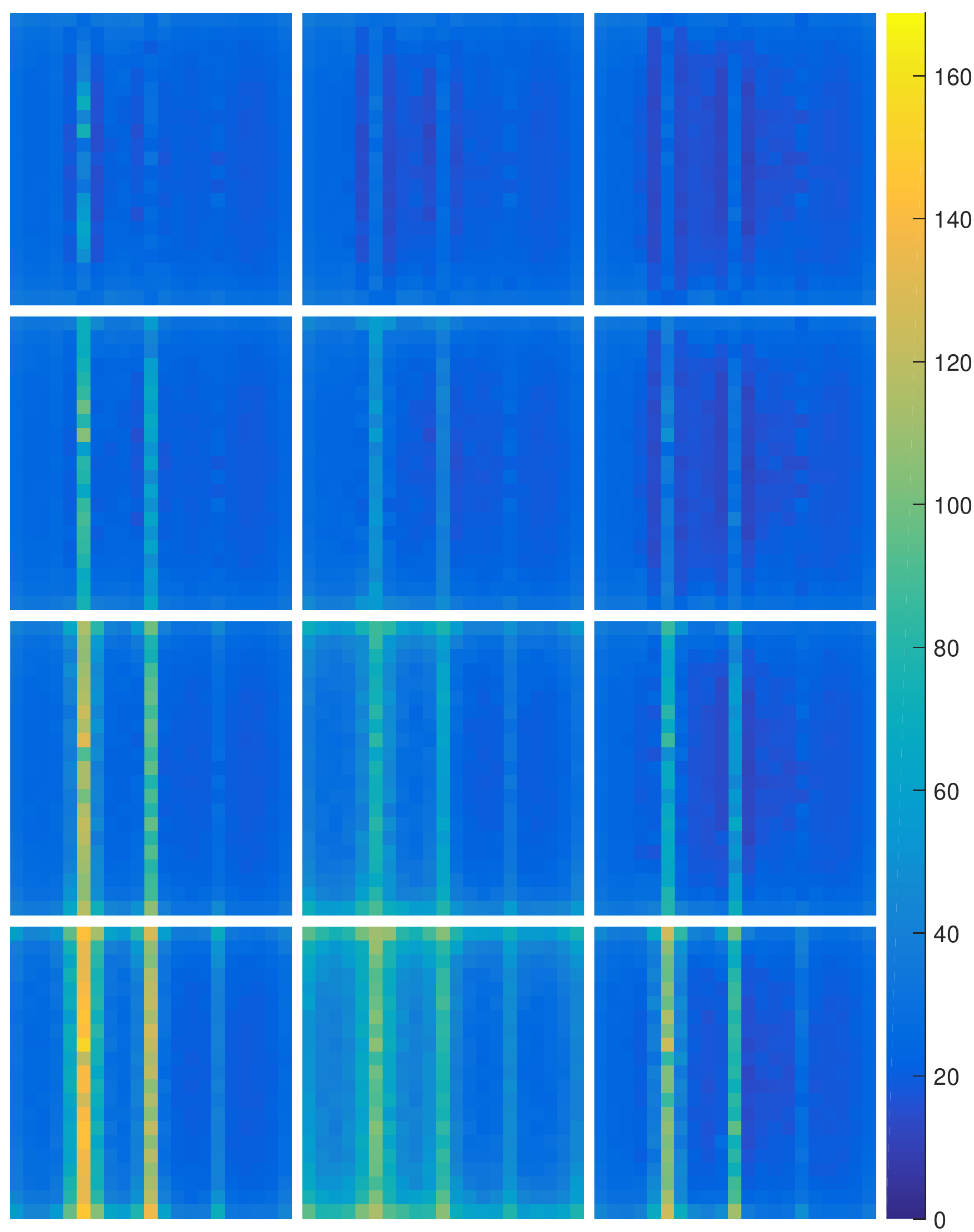}
\raisebox{13.6em}[0em][0em]{\hspace{-.25em}{\sf\tiny{{monocular\hspace{1.5em}stereo\hspace{1.4em}multispectral\hspace{.4em}}}}}\\
\raisebox{2.7em}[0em][0em]{\hspace{-10.6em}{\sf\tiny{\rotatebox{90}{{$\sigma=0.07$}\hspace{.35em}{$\sigma=0.05$}\hspace{.35em}{$\sigma=0.03$}\hspace{.35em}{$\sigma=0.01$}}}}}\\
\raisebox{10.5em}[0em][0em]{\hspace{9.5em}{\sf\tiny{\rotatebox{-90}{g/litre}}}}\\
\vspace{-3em}
\end{minipage}\vspace{-.5em} {
\caption{L) The absolute concentration estimation error for \HbO\ and \Hb\ combined across both test cases, showing mean absolute error and one standard deviation. For reference typical total haemoglobin concentration for whole blood in an adult male is approx. 145g/litre. Noise $\sigma$ is shown generally as for floating point image data, in a typical 8 bit sensor $\sigma=0.09$ corresponds to $\sigma_\text{8bit}= 23.04$. R) The mean concentration estimation error for monocular, stereo and multispectral at four noise levels. \label{f:noiseresults}}}
\vspace{-1.8em}
\end{figure}

We used the mesh-based MC (MMC) framework of \cite{MMCFang2010} with the digital phantom model shown in Fig. \ref{f:setup}.
For the MMC simulation photons were generated at intervals of 10nm across the range 400nm to 900nm. 
To detect the backscattered light photon momentum was recorded for all photons leaving the bounds of the meshed region. Photons that did not exit through the side of the mesh that was illuminated were discarded as were photons leaving at angles to the surface too oblique to be detected by a detector placed at 10mm away from the illuminated surface.
To simulate multispectral camera images of the scene the photons arriving at the detector were filtered into spectral bands. RGB images were generated by filtering the multispectral data with the response curves corresponding to RGB cameras, a stereo response curve is inset in Fig. \ref{f:setup}.
Noise was added to the multispectral data by adding zero mean normally distributed vales to each channel, for the RGB noise was generated correlated based on the response curve of the camera.

The performance of our method as seen in Fig. \ref{f:noiseresults} is close to the estimation from full multispectral data as the noise level increases. Also at low noise levels the stereo (6 channel) version of our method outperforms the monocular (3 channel) version. However at high levels of noise the stereo version underperforms due to the increased likelihood of over or under saturated pixel data in the six channels compared to the three of the monocular version. The impact of over or under saturated measurement data is more significant in our method compared to a multispectral approach because each channel in our method corresponds to wide wavelength range, which in the presence of the smoothness prior on the $I_\lambda$ causes large global under or over chromophore concentration estimation. The presence of a few saturated outliers has less impact on the multispectral method as it is directly fitting against the multispectral data, and the effect of a saturated outlier is localised to an individual wavelength band.\vspace{-.5em}

\subsection{In vivo validation}\vspace{-.5em}

Multispectral data from a porcine study was used to create a ground truth \Hb\ and \HbO\ concentration maps and corresponding RGB images. Multispectral data sets $M$ comprised 24 non-overlapping 10nm wide band limited images over the spectral range 460nm to 690nm. Multispectral haemoglobin estimation using the method of \cite{Clancy12} was performed on these data to establish a best case ground truth and the coefficient of determination (CoD) of this fit was calculated. We masked a subset $\hat{M}$ of the original multispectral data where the CoD was over $0.5$ creating a multispectral data set where we have high confidence in the ground truth concentration. From the 58 multispectral data sets we generated corresponding RGB images from typical RGB camera response curves. 
\begin{figure}[t]\vspace{-.6em}\hspace{-1em}
\begin{minipage}{0.545\linewidth}\centering
\includegraphics[width=\textwidth]{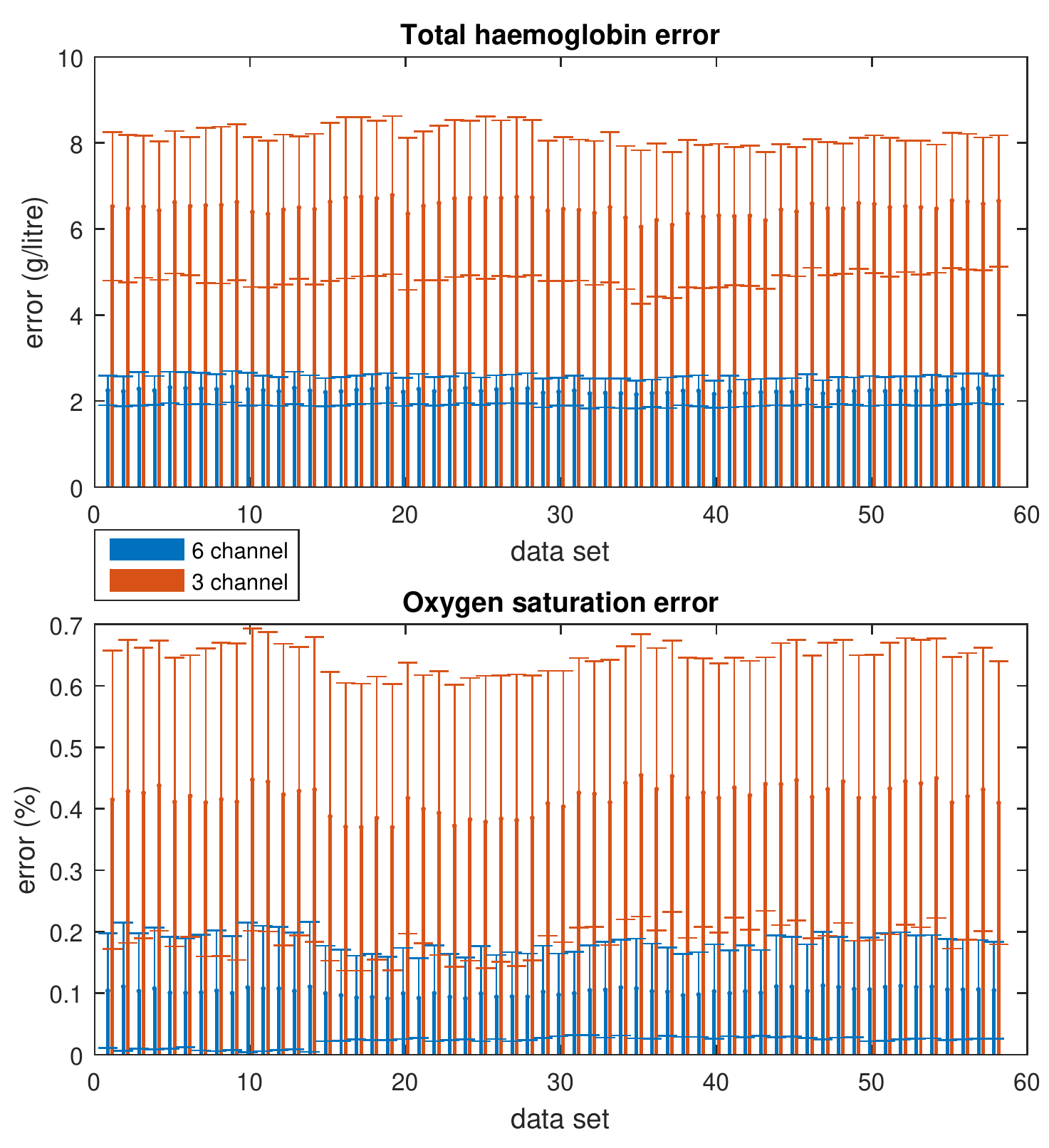}
\end{minipage}\hspace{-.9em}\hfill
\begin{minipage}{0.45\linewidth}\centering\hspace{-.em}\vspace{.1em}
\includegraphics[width=\textwidth]{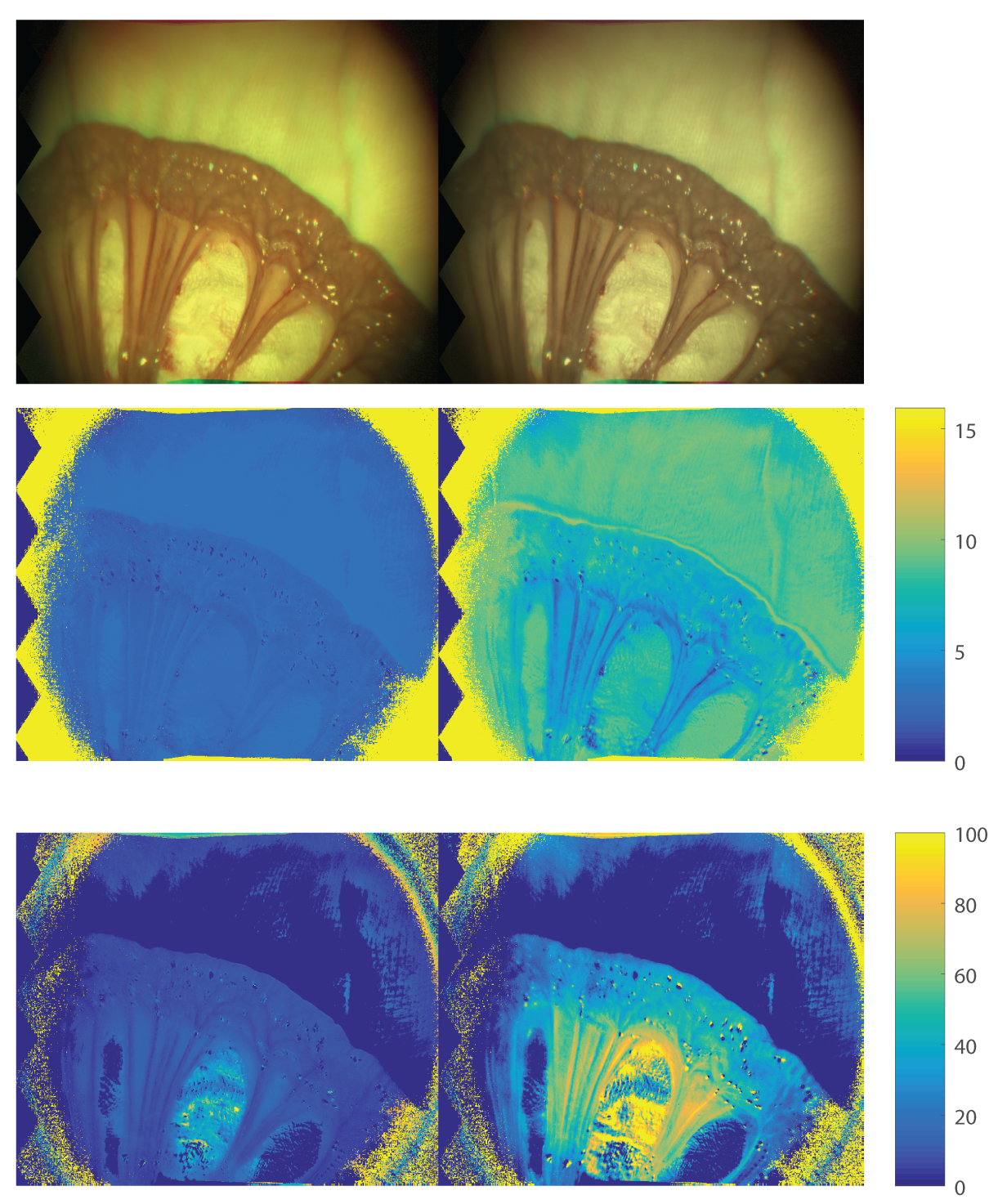}\\\raggedright
\raisebox{16.8em}[0em][0em]{\hspace{-.25em}{\sf\tiny{\rotatebox{90}{stereo rgb}}}}\\
\raisebox{9.05em}[0em][0em]{{\sf\tiny{\hspace{2.8em}{6 channel (stereo)}\hspace{4.6em} {3 channel (mono)}}}}\\
\raisebox{9.1em}[0em][0em]{{\sf\tiny{}}}\\
\raisebox{15.8em}[0em][0em]{\hspace{16.4em}\sf\tiny\rotatebox{-90}{g/litre}}\\
\raisebox{9.6em}[0em][0em]{\hspace{16.4em}\sf\tiny\rotatebox{-90}{\%}}\\
\raisebox{15.4em}[0em][0em]{\hspace{-0.25em}\sf\tiny\rotatebox{90}{total haemoglobin}}\\
\raisebox{10.5em}[0em][0em]{\hspace{-0.25em}\sf\tiny\rotatebox{90}{saturation}}\vspace{-7.3em}
\end{minipage}\vspace{-1.3em}
\caption{L) Estimation error for each of the 58 data sets comparing stereo (blue) and monocular (red) against multispectral derived ground truth. Clearly visible is the improved precision of the stereo method especially for the correct evaluation of the oxygen saturation. R) A registered stereo view of a surgical site with. Maps of the estimation error for saturation and total haemoglobin computed from stereo and monocular data.\label{f:algcomp}}\vspace{-1.8em}
\end{figure}

We ran both the monocular and stereo version of our method on $\hat{M}$ and comparison was then made against the multispectral derived the ground truth. Concentrations of \Hb\ and \HbO\  were then converted into the total haemoglobin and oxygen saturation measures since these are the markers that would then be used to clinical evaluate a surgical site. As shown in (\ref{f:algcomp}) the use of stereo significantly improves on the estimation of total haemoglobin with an overall mean absolute error of less then 3g/litre when using 6 channels from two cameras compared to over 6g/litre for a 3 channel monocular approach. The standard deviation of the error for each method remains close yet is slightly lower for the 6 channel variant. Given that total haemoglobin concentration of whole blood is in the region of 145g/litre this indicates a high degree of accuracy when imaging for the purposes of perfusion mapping. Oxygen saturation estimation shows the most marked improvement when using stereo over monocular with an overall saturation estimation error of 10.27\% down from 41.71\% for the stereo and mono variants respectively. In evaluating the {\em in vivo} performance, results that corresponded to a \THb\ concentration greater than 200g/litre were considered outliers this enumerated as less than 0.1\% of the results being rejected as outliers.

In both cases our method produces highly similar results to those from multispectral inference, and the error is typically located in areas not corresponding to vasculature. This is illustrated in Fig. \ref{f:algcomp} where a section of bowel is exposed on a gauze background. For the stereo case the error in \THb\ estimation is very low across the view however the \SatO\ estimation performs less well in areas of low \THb . This is to be expected as the oxygen saturation is a ratio of \Hb\ and \HbO\ and when both are at low concentration small errors in estimation of either in either become amplified in the aggregate saturation measure.
\vspace{-1.em}
\section{Discussion}\vspace{-.5em}

We have presented a novel estimation tool for measuring the concentration of \Hb\ and \HbO\ directly from laparoscopic RGB video. The method provides greater accuracy when applied to stereoscopic data as typically found in robotic assisted MIS. We have shown that the method performs well on synthetic data and is comparable to the result from raw MSI data acquired using modified imaging hardware such as a LCTF camera. Our method's only requirement is to have a calibration of the laparoscopic sensors and light source to capture the response curve of each channel. This makes our technique very applicable to a wide range of MIS procedures and easily to integrate in the operating theatre. The success of our stereo method is going to be strongly linked to the quality of the registration of the two camera views, while this remains an open problem there exist effective techniques specifically targeted at registration for multispectral inference \cite{Du2015} \cite{Clancy12}. Our method also requires imaging to be at a constant distance from the tissue surface, integrating the stereo acquired depth information may provide was to normalise for these global changes in irradiance.
%
%We have presented a novel single shot estimation tool for measuring the concentration of oxy and deoxy haemoglobin haemoglobin directly from the RGB video data utilising stereo views to provide greater accuracy. We have shown how our method comes close to the accuracy of that achieved from estimating haemoglobin concentration from multispectral data with out the requirements for alternative imaging equipment such as cameras or lighting other than that which would already be in use during a typical laporoscopic procedure. Our method's only requirement is for a calibration of the camera and light system to be made to capture the response curve of each channel. Given this initial calibration our method is able to run at close to real time $\sim1$sec per stereo 1080p image pair, on an Intel Xeon v2 2.6 GHz processor. It should be possible to achieve further speedups by storing the decomposition of $(C^{T}C+ \Gamma^{T} \Gamma )$ from Equation \ref{e:solve} instead of calculating the inverse at every step since it is invariant, reducing the computation to a single matrix multiplication and one back projection using the decomposition per RGB tuple.
%\vspace{-1.em}
%\section{Acknowledgements}\vspace{-1.em}
%
%\doubleblind{Neil Clancy is supported by an Imperial College Junior Research Fellowship. This work was also supported by ERC grant 242991.} 

\bibliographystyle{splncs03}
\bibliography{mybib}
%\pagelimit{8}
\end{document}